\documentclass[11pt]{article}

\usepackage[final]{acl}

\usepackage{times}
\usepackage{latexsym}

\usepackage[T1]{fontenc}

\usepackage[utf8]{inputenc}

\usepackage{microtype}

\usepackage{inconsolata}

\usepackage{amsmath,amsfonts,bm}

\def\eqref#1{equation~\ref{#1}}

\def\1{\bm{1}}

\DeclareMathAlphabet{\mathsfit}{\encodingdefault}{\sfdefault}{m}{sl}
\SetMathAlphabet{\mathsfit}{bold}{\encodingdefault}{\sfdefault}{bx}{n}

\usepackage{graphicx}
\usepackage[linguistics]{forest}

\usepackage[font=small,labelfont=bf]{caption} 
\usepackage{multirow}  
\usepackage{booktabs}  

\usepackage{xcolor}
\usepackage{comment}
\usepackage{listings}

\title{
MADE: A Living Benchmark for Multi-Label Text Classification with Uncertainty Quantification of Medical Device Adverse Events}

\usepackage{xspace,textcase}
\DeclareRobustCommand{\dataset}{\NoCaseChange{\texttt{MADE}}\xspace}

\author{Raunak Agarwal, {\bf Markus Wenzel}, {\bf Simon Baur}, \\ {\bf Jonas Zimmer}, {\bf George Harvey}, {\bf Jackie Ma} \\
Department of Artificial Intelligence\\
Fraunhofer Heinrich Hertz Institute\\
Berlin, 10587, Germany
\\
  \small{
    \textbf{Correspondence:} \href{mailto:jackie.ma@hhi.fraunhofer.de}{jackie.ma@hhi.fraunhofer.de}
  }
}

\begin{document}
\maketitle
\begin{abstract}
Machine learning in high-stakes domains such as healthcare requires not only strong predictive performance but also reliable uncertainty quantification (UQ) to support human oversight. 
Multi-label text classification (MLTC) is a central task in this domain, yet remains challenging  due to label imbalances, dependencies, and combinatorial complexity.
Existing MLTC benchmarks are increasingly saturated and may be affected by training data contamination, making it difficult to distinguish genuine reasoning capabilities from memorization.
We introduce \texttt{MADE}, a living MLTC benchmark derived from \underline{m}edical device \underline{ad}verse \underline{e}vent reports and continuously updated with newly published reports to prevent contamination.
\texttt{MADE} features a long-tailed distribution of hierarchical labels and enables reproducible evaluation with strict temporal splits.
We establish baselines across more than 20 encoder- and decoder-only models under fine-tuning and few-shot settings (instruction-tuned/reasoning variants, local/API-accessible). 
We systematically assess entropy-/consistency-based and self-verbalized UQ methods.
Results show clear trade-offs: 
smaller discriminatively fine-tuned decoders achieve the strongest head-to-tail accuracy while maintaining competitive UQ;
generative fine-tuning delivers the most reliable UQ; 
large reasoning models improve performance on rare labels yet exhibit surprisingly weak UQ; and self-verbalized confidence is not a reliable proxy for uncertainty. Our work is publicly available at \url{https://hhi.fraunhofer.de/aml-demonstrator/made-benchmark}.

\end{abstract}
\section{Introduction}

\begin{figure*}[t]
  \centering
  \includegraphics[scale=0.5]{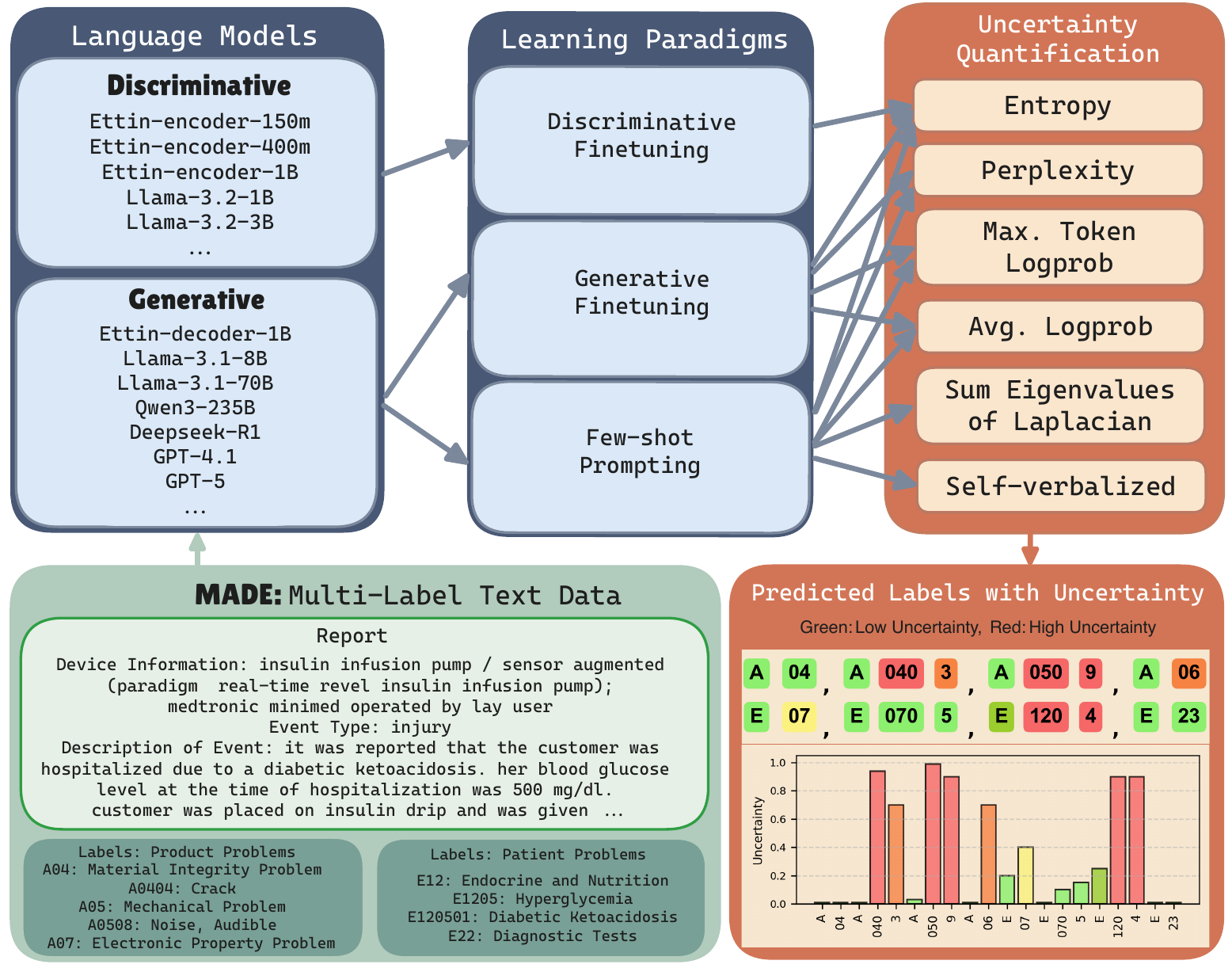}
  \caption{
  \textit{Top:} Overview of the benchmarking setup, encompassing discriminative and generative language models, 
  learning paradigms (discriminative or generative fine-tuning and few-shot prompting), and uncertainty quantification (UQ) approaches. 
  \textit{Bottom, left:} Multi-label text classification of medical device adverse events, each annotated with hierarchical product and patient problem labels.
  \textit{Bottom, right:} UQ quality is evaluated (for of each model, learning paradigm and UQ method).
  }
  \label{fig:overview}
\end{figure*}

Strong predictive performance is not sufficient for the adoption of machine learning (ML) models in high-stakes settings such as healthcare \citep{lekadir2025future, reddy2021evaluation} where human oversight is paramount \citep{Shneiderman02042020}. 
In this regard, uncertainty quantification (UQ) is essential for reliable ML systems \citep{ojha2025navigating} because it can flag doubtful or ambiguous cases for human re-examination.

Multi-label text classification (MLTC) is central to patient categorization, clinical coding, and incident reporting, etc. 
Developing reliable MLTC systems presents practitioners with several challenges: 
MLTC requires selecting multiple labels from a typically much larger set, which leads to a combinatorial problem that scales exponentially with the label space size. 
Real-world MLTC data can be characterized by severe inter- and intra-class imbalances: a few common conditions comprise the majority of examples, while safety-critical conditions reside in the long tail. 
Models must learn to disentangle correlated signatures without becoming biased toward frequent classes.
Further, labels often co-occur and are hierarchically interdependent, violating the assumption of label independence. 

Crucially, existing MLTC benchmarks are increasingly ill-suited for evaluating frontier large language models (LLMs). 
Traditional datasets are static, can be saturated, suffer from data contamination due to their inclusion in LLM pre-training corpora, or lack the label imbalance and interdependence of real-world environments.

These challenges leave practitioners with open questions:
Which model architecture is most suitable for this complexity? 
Can a specialized, small model solve the task, or is a larger foundation model required?
Which learning paradigm (fine-tuning vs.~in-context learning) yields the best trade-off between performance for frequent and rare classes?
How reliable are the resulting predictions? We address these questions by introducing an uncontaminated benchmark and systematically evaluating models across architectures, learning paradigms, and UQ methods.

\subsection{Related work}
\textbf{Multi-label text classification.} Early approaches to MLTC relied on models like Na\"ive Bayes, support vector machines \citep{wang-manning-2012-baselines}, or recurrent neural networks. 
With the advent of transformer-based language models \citep{vaswani2017attention}, encoder-only \citep{huang-etal-2021-balancing} and decoder-only \citep{ma2025largelanguagemodelsmultilabel,galke2025reallymakingprogresstext} architectures served for this task.
Controlled comparisons between generative and discriminative fine-tuning for decoder models and
head-to-head comparisons between encoder- and decoder-only models under matched conditions are  rare. 
\citet{weller2025seq} addresses this with a suite of matched encoder- and decoder-only models (`Ettin') trained on the ModernBERT architecture \citep{warner2024smarter} across model sizes, 
but do not evaluate on MLTC, leaving this comparison open. 
Decoder-only LLMs support zero-shot (inference without labeled data) and few-shot prompting (in-context learning; conditioning on a few 
labeled exemplars without updating model weights), which has been compared with fine-tuning on a variety of tasks where medical expertise is required, with differing results \cite[e.g.,][]{nori2023can, maharjan2024openmedlm, labrak-etal-2024-biomistral}.
Across broader benchmarks, the advantage of fine-tuning vs.~prompting varies by task, data regime, and model \citep{chen2025benchmarking,wu2025bridgebenchmarkinglargelanguage}. 
For MLTC, performance is sensitive to e.g. label semantics, output constraints, and thresholding strategies, motivating controlled comparisons.

\textbf{Benchmarking datasets} for multi-label text classification span diverse domains, 
including 
scientific literature (\citealp{kowsari2017hdltex}, 
\citealp{yang2018sgm}, 
\citealp{fallah2022adapting}, 
\citealp{chen2022multi}; 
\citealp{schopf2023exploring}),
newswire \citep{lewis_reuters21578}, 
finance-related user posts \citep{maia2021comparative}, 
patents \citep{tang2020multi}, 
clinical notes for ICD coding \citep{johnson2016mimic, mullenbach2018explainable}, 
legislative documents (\citealp{steinberger-etal-2012-jrc}, 
\citealp{boella2013system}, 
\citealp{chalkidis019large}, 
\citealp{bocchi-etal-2024-kevlar}), 
and more. 
Overlap between benchmark content and LLM pre-training corpora raises concerns about contamination that can inflate zero-/few-shot performance \cite{jacovi2023stopuploadingtestdata, oren2024proving, zhu-etal-2024-clean, li2024latesteval, deng-etal-2024-investigating, xu2024benchmarkingbenchmarkleakagelarge}. 
Heavy reuse can lead to benchmark saturation via overfitting and implicit adaptation,
which motivates the continuous introduction of novel datasets to assess generalization.
Typical benchmarks are scraped from online repositories and rely on distant or weak supervision (e.g., tags/metadata as labels), often with limited documentation, introducing label noise and hampering reproducibility. Multiple sources, pre-processing choices, copyright limitations, and dataset versions further complicate fair comparisons.


\textbf{Uncertainty quantification (UQ).}
For discriminative models, entropy-based UQ measures are well understood \citep{sensoy2021misclassification,mucsanyi2024benchmarking,baur2025benchmarking}.
In contrast, UQ for generative models is more complex: token-level probabilities, consistency across stochastic generations, and self-verbalized confidence introduce distinct challenges. However, estimating uncertainty in LLM benchmarks is of great interest \citep{bean2025measuringmattersconstructvalidity} and
recent works \cite[e.g.,][]{ye2024benchmarking, vashurin-etal-2025-benchmarking} cover different natural language processing tasks, but not yet MLTC.
Token- or information-based uncertainty metrics, which in various forms rely on the log-probabilities of an LLM's outputs, have been surveyed and formalized by \citealp{shorinwa2025survey,fomicheva2020unsupervised}. 
Among those of interest are: entropy of the top-$n$ log-probabilities, perplexity, average or maximum token log-probability.
Consistency-based UQ metrics measure the output variability of a model under stochasticity, and have been widely used for LLMs as a way to capture epistemic uncertainty \citep{xiao2025consistency}. 
\citealp{lin2024generating} suggested an effective approach to derive a consistency-based uncertainty score through the sum of eigenvalues of a graph Laplacian. 
\citealp{vashurin2025uncertainty} proposed the combination of information- and consistency-based metrics as an effective strategy to improve uncertainty estimation, motivating our approach of 
integrating token-level uncertainty with consistency measures.
Self-verbalized uncertainty, in which a model outputs a confidence score corresponding to its prediction \cite{kadavath2022languagemodelsmostlyknow, tian-etal-2023-just, harsha2024quantifying}, offers a complementary approach to directly capture model-reported uncertainty, especially when log-probabilities are not available.
\citealp{dong2025survey} review confidence calibration for imbalanced data.

We focus on information-based and consistency-based metrics due to their complementary strengths, and additionally test self-verbalized uncertainty given its practical appeal when log-probabilities are unavailable.

\subsection{Our contributions} 
In this paper, we
(1) \textbf{create a benchmark:} 
To counteract saturated benchmarks with potential pre-training exposure, 
we introduce a reproducible pipeline to generate a challenging evaluation suite based on medical device adverse event reports made available by the U.S.~Food and Drug Administration (FDA).
We release \dataset, 
characterized by a long-tailed distribution of 1,154 interdependent labels across three hierarchical levels, specifically curated to test generalization limits.
We dub this MLTC dataset as a `living benchmark' because the continuous publication of new reports by the FDA will continue to enable testing of models 
on fresh data, avoiding potential leakage of test data into the pretraining corpora of future foundation models.
(2) \textbf{Establish solid baselines:} We fine-tune more than twenty encoder- and decoder-only models in discriminative and generative settings and compare them with few-shot prompting of local and API-accessible models (including reasoning variants).
(3) \textbf{Evaluate UQ capabilities:}
We systematically study multiple UQ approaches (see Figure~\ref{fig:overview}) --- including 
information-based, consistency-based, and self-verbalized uncertainty---and assess their utility for sample routing or human-in-the-loop triage. Beyond providing practical guidance on model selection and UQ strategies for MLTC, our results expose critical reliability trade-offs.
By releasing \dataset alongside comprehensive baselines, we share a foundation for future research into reliable MLTC, 
enabling the community
to evaluate the limits of increasingly capable language models.

\section{Data: Medical Device Adverse Events}
\label{sec:data}

The FDA regularly publishes medical device adverse event reports\footnote{\url{https://open.fda.gov}} 
along with annotations that capture the associated product and patient problems.
From this resource, we curate a large-scale text classification dataset with hierarchical multi-labels.
Statistics of the dataset, hierarchical labels, and topics reported are provided in Table~\ref{tab:dataset_stats}, Figure~\ref{fig:frequent_problems}, and Figure~\ref{fig:topics} (in Appendix~\ref{subsec:additional_results}).
We collect files from 2015 (previous reports lack the necessary product problem labels) until mid-2025. 
From each report, we extract the event description, relevant metadata (event type and device information), and the product and patient problem labels.

While FDA coders assign the IMDRF labels, the consistency of this annotation process has not been formally assessed through inter-annotator agreement studies. Labels may therefore reflect annotator variability or systematic biases in the reporting pipeline. As operational annotations, they may not meet the standards of clinically validated ground truth.

\begin{figure*}[!h]
\centering
\includegraphics[width=2\columnwidth]{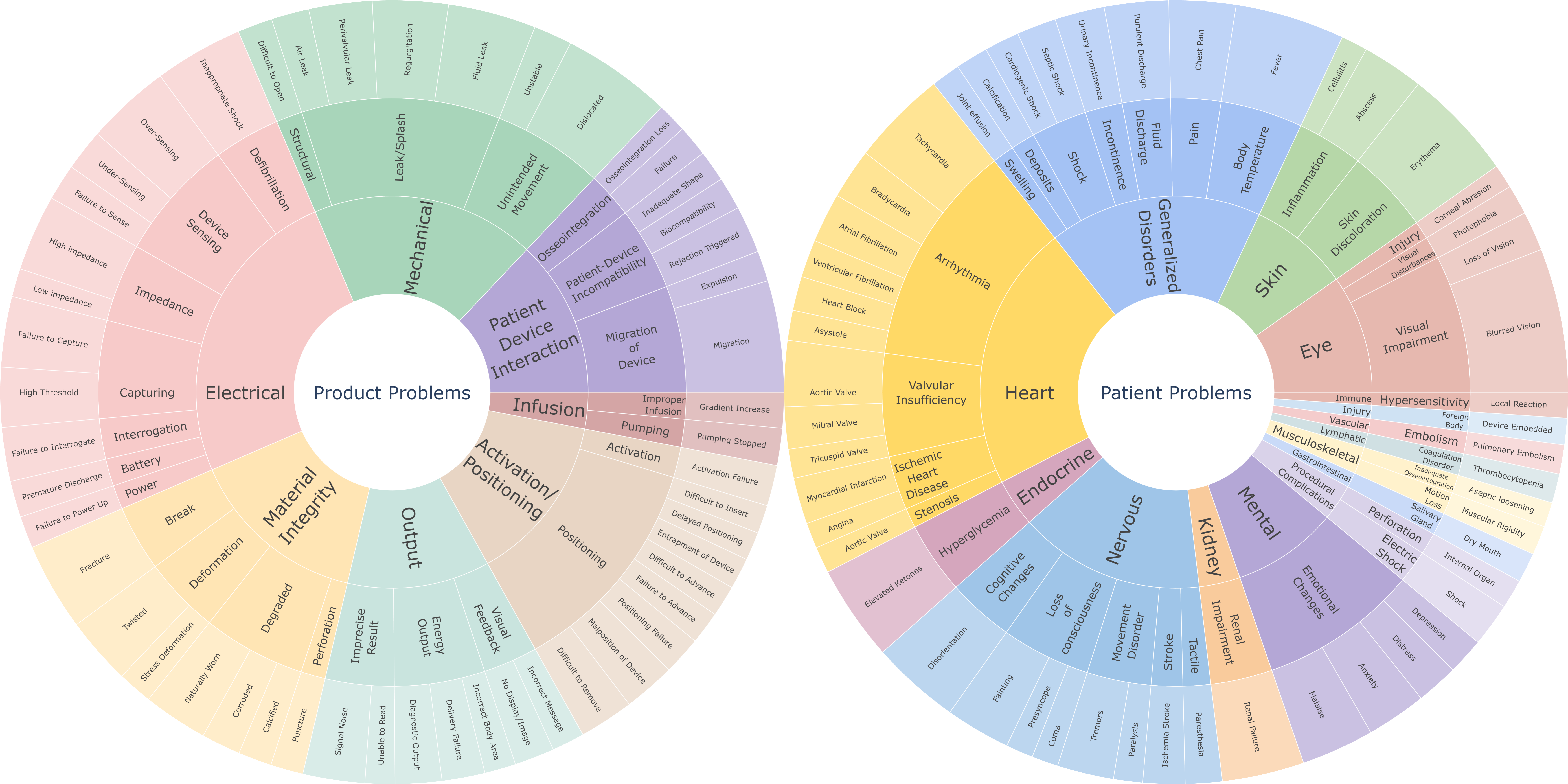}  
\caption{Product and patient problems are the hierarchical multi-labels of \dataset.
The outer ring shows the fifty most frequent product or patient problems in the test set, grouped by their parent classes (middle ring) and grandparent classes (inner ring).}
\label{fig:frequent_problems}
\end{figure*} 

\begin{table}[]
\footnotesize
\centering
\begin{tabular}{lr}
\toprule
Total number of samples & 488,273 \\
\quad Training set \textit{(2015--2023)} & 298,825 \\
\quad Validation set \textit{(1--6/2024)} & 71,271 \\
\quad Test set \textit{(7/2024 -- 6/2025)} & 118,177 \\
\quad Truncated test set & 10,288 \\
\midrule
Average tokens (cl100k\_base) & $\sim$370 \\
Average labels per sample & 8.79 \\
\midrule
Unique labels & 1,154 \\
Hierarchy levels of labels & 3 \\
Minimum occurrences per label & 5 \\
\bottomrule
\end{tabular}
\caption{Summary statistics of \dataset.}
\label{tab:dataset_stats}
\end{table}


\begin{figure}[h!]
\includegraphics[width=\columnwidth]{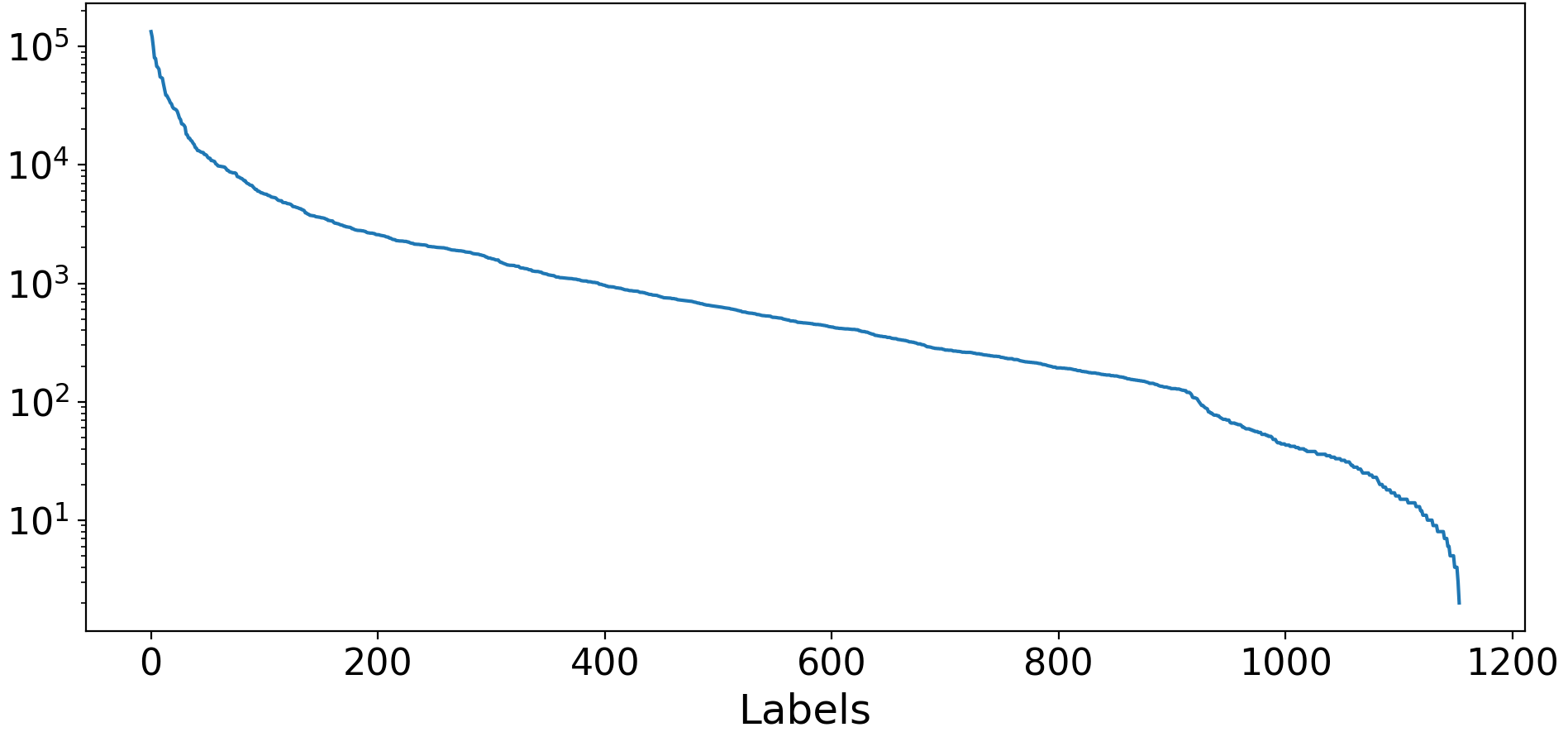}
\caption{Log-scaled distribution of label frequencies illustrating severe imbalance and the pronounced long-tail pattern.}
\label{fig:distribution_of_label_frequencies}
\end{figure}

We map these labels from FDA terms for each product and patient problem (see Figure~\ref{fig:frequent_problems}) to hierarchical codes provided by the International Medical Device Regulatory Forum~\citep{IMDRF2025AdverseEvent}.
Samples missing IMDRF codes for one or more terms are excluded. 
To leverage the hierarchy of the IMDRF terminology, each code is up-propagated to include all ancestor codes, as determined from official IMDRF annexes. 
This approach recognizes that FDA annotators typically use the most specific (leaf) codes and ensures that models are not penalized for predicting valid parent codes.
Label mapping and propagation yield two sets of hierarchical labels per report, which are flattened into a union of target labels. 
To avoid introducing unseen labels into the test set, we freeze the label taxonomy at December 2023, and discard labels introduced thereafter.
We remove extremely rare labels ($<$5 instances), yielding a final set of 1,154 labels following a long-tail distribution. 
(see Figure~\ref{fig:distribution_of_label_frequencies}).

We deduplicate reports based on the event description, retaining only the first occurrence, thus reducing the dataset to 5.5 million samples,
and further downsample by applying HDBSCAN \citep{McInnes2017} on embeddings of event descriptions, selecting cluster representatives, and retaining a high proportion of rarer labels; all events labeled death are included.
Data from 2015 to 2023 serve for training, the first half of 2024 for testing, and July 2024 to June 2025 for testing. 
We compare these periods with the models' knowledge cutoff dates in Appendix~\ref{subsec:knowledge_cutoff_dates}.
To limit inference costs while maintaining statistical validity, we create a representative `truncated test set' via stratified sampling (cf.~Table~\ref{tab:dataset_stats}) and perform all evaluations on this split.
In summary, our pipeline transforms raw FDA adverse event reports into an IMDRF-compliant, hierarchically-labeled, temporally-split dataset tailored for a challenging MLTC benchmark. Because the FDA releases new reports quarterly, researchers can evaluate future models on data guaranteed to post-date their training, providing ongoing contamination-free evaluation rather than a one-time static benchmark.

See Appendix \ref{subsec:data_availability} for data availability.

\section{Experimental setup}

\subsection{Fine-tuning and in-context learning}

\textbf{Discriminative training.} 
We fine-tune Llama 3.2-1B/-3B, and 3.1-8B base models  \citep{grattafiori2024llama3herdmodels}, with a classification head in the final layer.
We also fine-tune Ettin models \citep[150M, 400M, 1B; ][]{weller2025seq}, which are based on the ModernBERT \citep{warner2024smarter} architecture. 

We experiment with a hierarchical loss where the label space is partitioned by taxonomy level and a separate binary cross-entropy loss is computed at each level, with the total loss being their unweighted sum. Details are reported in the Appendix~\ref{subsec:additional_results}. 

We use AdamW with cosine scheduler, batch size 512, 20 epochs, warmup ratio 0.1, context length 512, and fixed maximum gradient norm 0.01. Learning rate is tuned in [2e-4, 5e-4]. Per-label classification thresholds are selected to maximize F1 on the validation set independently for each label.

\textbf{Generative training.} 
We fine-tune Llama 3.2-1B, 3.2-3B, 3.1-8B and 3.1-70B, and Ettin decoder variants (400M, 1B) to generate class labels as tokens. Each model is trained for 4 epochs. An improvement for a larger number of epochs was not observed. We compare full fine-tuning with LoRA \citep{hu2022lora} for selected models. 

We use AdamW~\cite{loshchilov2019decoupledweightdecayregularization} 
with cosine learning-rate decay~\cite{Loshchilov2016SGDR}, warmup ratio 0.1, batch size 64, context length 512, and maximum generation length 50 tokens. Learning rate (3e-5 to 2e-4) and maximum gradient norm (1.0--10.0) are tuned on the validation set.
For LoRA experiments, we use rank 16--64 and alpha 32--64, targeting all attention (Q, K, V, O) and MLP (gate, up, down) projection layers.

\textbf{Few-shot prompting/in-context learning.}
We locally host Llama 3.2-3B, 3.1-8B, and 3.1-70B, DeepSeek-R1
\citep[37B active, 671B total; ][]{deepseekai2025deepseekr1incentivizingreasoningcapability}, Qwen3 \citep[4B, 30B, 235B; ][]{yang2025qwen3technicalreport}, Kimi K2 \citep[32B activated; 1T total; ][]{kimiteam2025kimik2openagentic}, gpt-oss-120b \citep[5.1B active, 117B total; ][]{openai2025gptoss120bgptoss20bmodel}, GLM-4.5-Air
\citep[12B active, 106B total; ][]{glmteam2025glm45agenticreasoningcoding}, and Llama-3.3-Nemotron-49B-v1.5 \citep[][]{bercovich2025llamanemotronefficientreasoningmodels}.
GPT-4.1 \citep[][]{openai2025introducinggpt41} and GPT-5 \citep[][]{openai2025gpt5} are accessed via API. 
Among API-accessed models, only GPT-4.1 -- but not GPT-5 -- provides log probabilities, which are required for our UQ investigations. GPT-5 is run with medium reasoning effort and does not allow modifying temperature. Knowledge cutoff dates for all the models are discussed in Appendix~\ref{subsec:knowledge_cutoff_dates}. 
The prompt includes task instructions, the label list, and ten kNN-retrieved training examples
(see Appendix~\ref{subsec:Prompting_setup}). A
pre-trained model is used to embed the reports for retrieval \citep[`bioclinical-modernbert-base-embeddings', ][]{neuml_bioclinical_modernbert_base_embeddings}. Each retrieved example is truncated to 10000 characters (about 2300 tokens). Unless otherwise stated, all outputs are generated in a greedy decoding setup ($temperature = 0, top\_p = 1, top\_k = -1$).

\textbf{Software.} All experiments use PyTorch 2.7.1, Transformers 4.56.0, and vLLM 0.10.1. 

\textbf{Hardware.} Details about the computing infrastructure are reported in Appendix~\ref{subsec:computing_infrastructure}.

\subsection{Methods for uncertainty quantification}
\label{sec:uncertainty_quantification_methods}
For discriminative models, uncertainty is computed from the output probability distributions with per-label entropy across classes.
For generative models, we quantify uncertainty with two complementary approaches: (i) information-level uncertainty $U_{\text{info}}$ is measured with token-based metrics (entropy, improbability, avg-log-probability, perplexity), while (ii) consistency-based uncertainty $U_{\text{cons}}$ measures variation across multiple stochastic forward passes via the sum of eigenvalues of the graph Laplacian ~\cite{lin2024generating}.
Multiplying $U_{\text{info}}$ and $U_{\text{cons}}$ yields the combined uncertainty $U_{\text{combined}}$. 
We obtain self-verbalized uncertainty $U_{\text{self}}$ by prompting the model to express its own confidence. 
Appendix \ref{sec:uncertainty_quantification_methods_appendix} contains full metric definitions and details.

\subsection{Evaluation of uncertainty scoring}
\label{sec:uncertainty_quantification_evaluation}
We assess UQ quality 
using selective prediction.
Prediction Rejection Rate \citep[PRR, ][]{malinin2021uncertainty, vashurin2025uncertainty} quantifies the effectiveness with which uncertainty scores identify unreliable predictions. Model outputs are ranked according to the uncertainty scoring. Predictions with the highest uncertainty are progressively removed, starting from the most uncertain. The improvement of the overall correctness due to this rejection is measured with the Jaccard score. Performance gains (relative to a random ranking) are compared against an oracle ranking, which represents the theoretical upper bound. Figure~\ref{fig:prr_illustration} illustrates the intuition of PRR and how it is derived.
As complementary measure, we compute the Spearman correlation ($\rho$) between uncertainty scores and per-sample correctness.
A negative correlation indicates that a higher uncertainty aligns with a lower correctness, which is desirable.
A positive correlation suggests that incorrect predictions often occur with high confidence, which is unfavorable. 

Finally, we evaluate calibration using positive-class expected calibration error ($\text{ECE}{_+}$), a variant of expected calibration error \cite[ECE, ][]{lichtenstein1977calibration,dawid1985calibration} computed only on positive instances. Standard ECE is unreliable in our setting due to severe label imbalance and the multi-label structure of our data: for many medium, tail, and extreme-tail classes, it yields values below 1\% despite frequent missed predictions. This issue is exacerbated for LLM-based models, where log-probabilities are available only for positive predictions, requiring zero confidence to be assumed for negative predictions and artificially inflating calibration scores for rare labels. Conditioning ECE on class labels has been proposed to address shortcomings of ECE \cite{nixon2019measuring}. Therefore, we report $\text{ECE}_{+}$ (Appendix~\ref{sec:uncertainty_quantification_evaluation_appendix}), which can be interpreted as a measure of under-confidence.

\section{Results and Discussion}

\begin{table*}[h!]
\centering
\footnotesize
\begin{tabular}{@{}l@{}c@{}c@{}c@{}c@{}c@{}c@{}c@{}c@{}c@{}}
\toprule
\textbf{Paradigm}/model & \multicolumn{5}{@{}c@{}}{\textbf{Macro F1 $\uparrow$}} & \textbf{J $\uparrow$} & $\textbf{PRR} \uparrow$ & \textbf{$\boldsymbol{\rho} \downarrow$} & \textbf{ECE}$_{+} \downarrow$\\
\cmidrule(lr){2-6}
& Overall & Head & Medium & Tail & ET & & & & \\
\textit{Number of classes $\rightarrow$} & \textit{1154} & \textit{144} & \textit{481} & \textit{348} & \textit{181} & & & \\
\midrule


\textbf{Discriminative fine-tuning} & \ \textbf{0.51}$\pm$.03 \, & \ \textbf{0.72}$\pm$.02 \, & \ \textbf{0.62}$\pm$.03 \, & \ \textbf{0.5}$\pm$.03 \, & 0.12$\pm$.02 \, & \ \textbf{0.59}$\pm$.02 \, & 0.46$\pm$.05 \, & \ -0.40$\pm$.04 \, & \ 0.59$\pm$.03\\
\hline \vspace{-.2cm} \\
Llama-3.1-8B-Base & \underline{\textbf{0.54}} & \underline{\textbf{0.74}} & \underline{\textbf{0.64}} & \underline{\textbf{0.53}} & 0.12 & \underline{\textbf{0.62}} & 0.47 & -0.40 & 0.58\\
Llama-3.2-3B-Base & 0.51 & 0.72 & 0.62 & 0.49 & 0.11 & 0.59 & 0.46 & -0.41 & 0.59\\
Llama-3.2-1B-Base & 0.51 & 0.71 & 0.60 & 0.48 &\textbf{0.14} & 0.58 & \textbf{0.52} & \textbf{-0.42} & 0.60 \\
Ettin-1B-Encoder & 0.53 & 0.73 & 0.63 & 0.51 & 0.13 & 0.61 & 0.46 & -0.40 & \textbf{0.56} \\
Ettin-400m-Encoder & 0.51 & 0.72 & 0.61 & 0.50 & 0.12 & 0.58 & 0.44 & -0.36 & 0.59 \\
Ettin-150m-Encoder & 0.46 & 0.68 & 0.56 & 0.44 & 0.07 & 0.55 &  0.38 & -0.30 & 0.64 \\
\midrule

\textbf{Generative fine-tuning} & 0.48$\pm$.04 & 0.67$\pm$.03 & 0.57$\pm$.04 & 0.46$\pm$.04 & 0.11$\pm$.03 & 0.58$\pm$.06 & \textbf{0.57}$\pm$.03 & \textbf{-0.44}$\pm$.08 & \textbf{0.57}$\pm$.04\\
\hline \vspace{-.2cm} \\
Llama-3.1-70B-Base & \textbf{0.53} & \textbf{0.73} & \textbf{0.62} & \textbf{0.51} & \textbf{0.16} & \textbf{0.61} & 0.55 & -0.27 & \textbf{0.49} \\
Llama-3.1-8B-Base & 0.50 & 0.70 & 0.59 & 0.48 & 0.12 & 0.59 & \underline{\textbf{0.63}} & -0.30 & 0.52 \\
Llama-3.2-3B-Base & 0.48 & 0.67 & 0.57 & 0.46 & 0.12 & 0.58 & 0.60 & \underline{\textbf{-0.46}} & 0.57 \\
Llama-3.2-1B-Base & 0.43 & 0.63 & 0.52 & 0.39 & 0.10 & 0.45 & 0.54 & -0.44 & 0.60 \\
Ettin-1B-Decoder & 0.47 & 0.67 & 0.56 & 0.46 & 0.10 & 0.57 & 0.56 & -0.43 & 0.57 \\
Ettin-400m-Decoder & 0.44 & 0.66 & 0.54 & 0.42 & 0.07 & 0.54 & 0.57
 & -0.44 & 0.60 \\
\midrule

\textbf{Prompting -- instruct} & 0.29$\pm$.15 & 0.49$\pm$.16 & 0.35$\pm$.16 & 0.25$\pm$.16 & 0.08$\pm$.09 & 0.43$\pm$.17 & 0.49$\pm$.15 & -0.15$\pm$.23 & 0.68$\pm$.14\\
\hline \vspace{-.2cm} \\
Llama-3.1-70B-Instruct & 0.30 & 0.50 & 0.35 & 0.25 & 0.08 & 0.43 & \textbf{0.60} & -0.15 & 0.68 \\
Llama-3.1-8B-Instruct & 0.08 & 0.28 & 0.09 & 0.03 & 0.01 & 0.22 & 0.20 & 0.26 & 0.78 \\
Qwen3-235B-A22B-Instruct & \textbf{0.44} & \textbf{0.60} & \textbf{0.48} & \textbf{0.42} & \textbf{0.24} & 0.49 & 0.56 & \textbf{-0.34} & \textbf{0.56} \\
Qwen3-30B-A3B-Instruct & 0.22 & 0.48 & 0.27 & 0.14 & 0.05 & 0.43 & 0.54 & 0.05 & 0.59 \\
Qwen3-4B-Instruct & 0.29 & 0.49 & 0.35 & 0.25 & 0.09 & 0.41 & 0.49 &  -0.27 & 0.68 \\
Kimi-K2-Instruct & 0.09 & 0.18 & 0.11 & 0.06 & 0.01 & 0.07 & 0.28 & 0.08 & 0.97 \\
GPT-4.1 & 0.43 & 0.59 & 0.47 & \textbf{0.42} & 0.22 & \textbf{0.57} & 0.45 & -0.31 & 0.60 \\
\midrule

\textbf{Prompting -- thinking} & 0.44$\pm$.05 & 0.58$\pm$.05 & 0.48$\pm$.05 & 0.42$\pm$.06 & \textbf{0.26}$\pm$.07 & 0.47$\pm$.04 & 0.21$\pm$.10 & -0.07$\pm$.04 & 0.59$\pm$.07 \\
\hline \vspace{-.2cm} \\
Llama-3.3-Nem.-49B-v1.5 & 0.42 & 0.57 & 0.46 & 0.38 & 0.19 & 0.46 & 0.21 & -0.03 & 0.59 \\
Qwen3-235B-A22B-Think. & 0.49 & 0.62 & 0.52 & 0.48 & 0.33 & 0.48 & \textbf{0.34} & \textbf{-0.09} & \underline{\textbf{0.45}} \\
Qwen3-30B-A3B-Think. & 0.45 & 0.58 & 0.49 & 0.44 & 0.28 & 0.47 & 0.08 & -0.07 & 0.56 \\
Qwen3-4B-Thinking & 0.38 & 0.53 & 0.42 & 0.36 & 0.2 & 0.43 & 0.21 &  -0.02 & 0.63 \\
DeepSeek-R1-0528 & 0.48 & 0.62 & 0.51 & 0.47 & 0.30 & 0.50 & 0.24 & \textbf{-0.09} & 0.50 \\
GLM-4.5-Air & 0.42 & 0.56 & 0.46 & 0.39 & 0.24 & 0.44 & 0.24 & \textbf{-0.09} & 0.62 \\
gpt-oss-120b & 0.40 & 0.57 & 0.45 & 0.38 & 0.15 & 0.45 & 0.05 & 0.00 & 0.63 \\
GPT-5 & \underline{\textbf{0.54}} & \textbf{0.68} & \textbf{0.58} & \underline{\textbf{0.53}} & \underline{\textbf{0.34}} & \textbf{0.57} & NA & NA & NA \\
\bottomrule
\end{tabular}

\caption{
Predictive performance (macro F1, J) and UQ (PRR, $\rho$, ECE$_{+}$) 
per learning paradigm and model.
Median results ($\pm\sigma$) are listed per paradigm, see also Figure~\ref{fig:comparison_discriminative_generative_prompt}).
Macro F1 is reported for head, medium, tail, and extreme-tail (ET) classes.
For discriminative fine-tuning, per-label thresholds were selected on the validation set.
For generative models, PRR corresponds to the respective best $U_{\text{info}}$ metric (see Section~\ref{sec:uncertainty_quantification_evaluation}).
Bold marks the best model within each paradigm; underlining indicates the overall best. 
Arrows show whether higher 
or lower 
values are preferable.
Evaluation used the truncated test set ($n = 10{,}288$).
Llama-3.1-8B-Base (discriminative) delivers the strongest predictive performance next to GPT-5, while Llama-3.2-3B-Base (generative) achieves the best UQ with respect to PRR and $\rho$, and Llama-3.2-1B-Base (discriminative) obtains the best ECE$_{+}$.
}
\label{tab:comparison_discriminative_generative_prompt}
\end{table*}

\begin{figure}[]
    \includegraphics[width=\columnwidth]{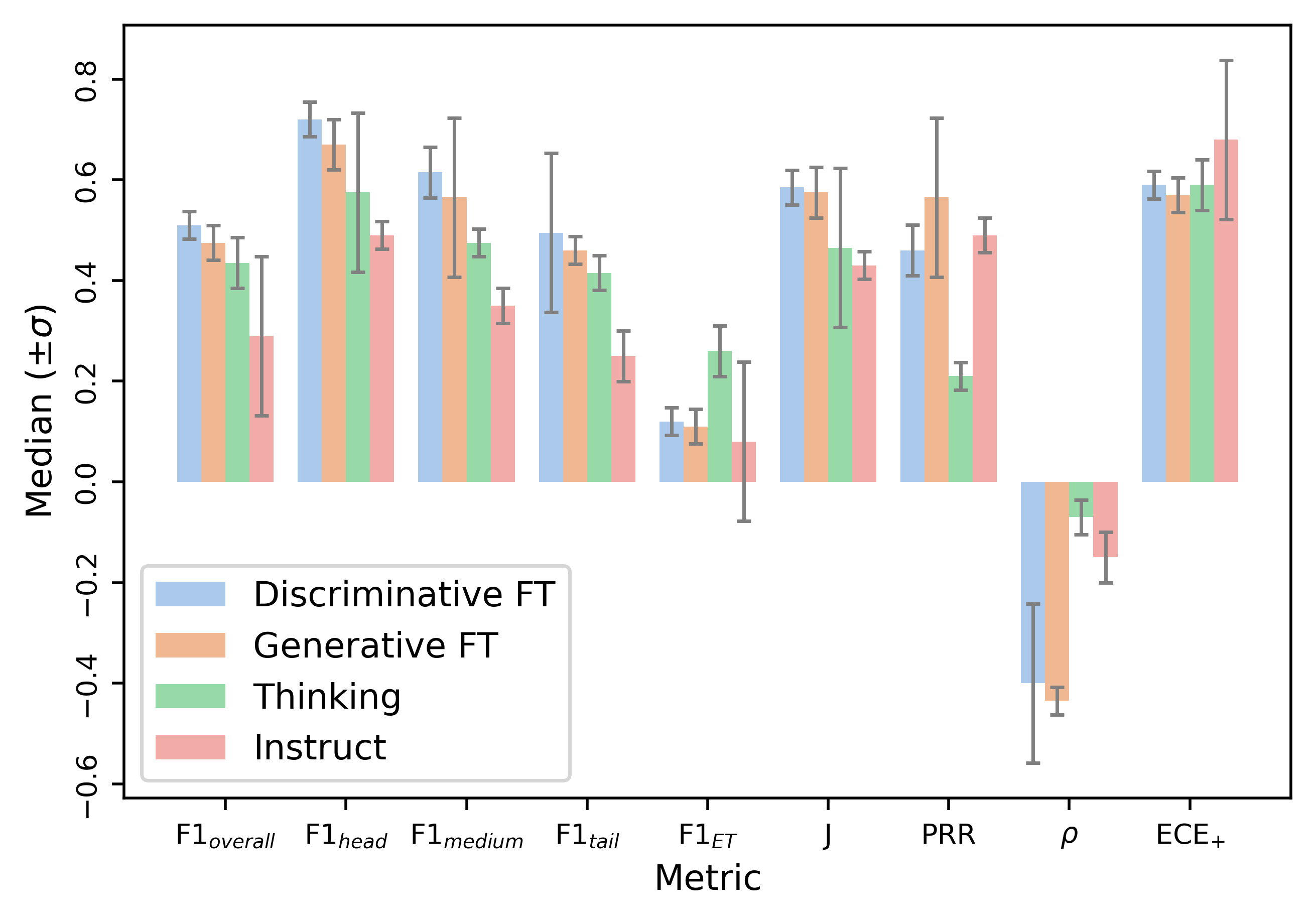}
    \caption{Overview of predictive performance and uncertainty quantification results per paradigm.}
    \label{fig:comparison_discriminative_generative_prompt}
\end{figure}

We evaluate predictive performance (macro F1, Jaccard $J$) and uncertainty quantification (PRR, Spearman $\rho$, $\text{ECE}{_+}$) 
across four paradigms: (i) discriminative fine-tuning of encoders/decoders, (ii) generative decoder fine-tuning, (iii) kNN-based ten-shot prompting with instruction-tuned or (iv) thinking/reasoning models. 
Labels are grouped by training-set frequency: head ($>$1\%), medium (0.1--1\%), tail (0.01--0.1\%), and extreme tail ($<$0.01\%).
Figure~\ref{fig:comparison_discriminative_generative_prompt} summarizes all results per paradigm with individual metrics shown in Table~\ref{tab:comparison_discriminative_generative_prompt}.

\subsection{Multi-label predictive performance}
\label{subsec:multi_label_predictive_performance}

\textbf{Fine-tuning.} Llama-3.1-8B-Base yields the highest macro F1 overall (0.54) and leads in head (0.74), medium (0.64), and tail (0.53) classes, when discriminatively fine-tuned. Smaller LLaMA models (Llama-3.2-3B/-1B) and encoders (Ettin-1B/400m/150m) trail slightly, consistent with their reduced capacity, but Ettin-1B-Encoder remains competitive given its size. In a generative setting, the much larger Llama-3.1-70B-Base achieves similar results. 
Notably, discriminative fine-tuning outperforms generative fine-tuning of similarly-sized decoders (Wilcoxon signed-rank test, $n=5$ matched pairs; $p\le0.05$).
Ablations of full vs. parameter-efficient fine-tuning and of instruction-tuned vs. base models are shown in Appendix~\ref{subsec:additional_results}. 

\textbf{Prompting.} For ten-shot in-context learning, Qwen3-235B-A22B-Instruct achieves the strongest macro F1 (overall 0.44, extreme tail 0.24), with only GPT-4.1 matching performance on tail classes. Reasoning models excel on the extreme tail (ET): GPT-5 attains the highest ET macro F1 (0.34) and ties the best overall macro F1 (0.54) with Llama-3.1-8B-Base (discriminative). Among open-weight reasoning models, Qwen3-235B-A22B-Thinking is strongest (overall 0.49, tail 0.48, ET 0.33). Gain on rare labels for all prompt-based models comes with a trade-off: head-class performance is consistently below that of the best fine-tuned decoders.


\textbf{Overall.} Discriminative decoder fine-tuning (Llama-3.1-8B-Base) remains the best for head--tail accuracy and consistently outperforms generative training. Prompt-based reasoning models (Qwen3-235B, DeepSeek-R1, GPT-5) dominate ET classes and can match overall F1 but lag on head classes (Mann-Whitney U test, independent groups: $n=9$ vs. $12$; $p\le0.05$). Within prompting, reasoning variants consistently outperform their instruct counterparts (see Qwen3).

\begin{table}[]
\centering
\small
\renewcommand{\arraystretch}{1.1}
\setlength{\tabcolsep}{3pt} 
\begin{tabular}{@{}l c c c c@{}}
\toprule
\textbf{$U_{\text{info}}$ metric} & \textbf{Gen. FT} \quad & & \textbf{Instruct} & \textbf{Thinking} \\
\midrule
Avg. Log-Prob. & \underline{\textbf{0.54}} $\pm$ 0.05 & & 0.37$\pm$0.25 & 0.18 $\pm$ 0.12 \\
Entropy & \underline{\textbf{0.58}} $\pm$ 0.03 & & 0.45 $\pm$0.15 & 0.19 $\pm$ 0.12 \\
Improbability & \underline{\textbf{0.54}} $\pm$ 0.05  & & 0.43 $\pm$ 0.15  & 0.17 $\pm$ 0.12  \\
Max. Log-Prob. & \underline{\textbf{0.52}} $\pm$ 0.08 & & 0.41 $\pm$ 0.16 & 0.18 $\pm$ 0.12 \\
Perplexity & \underline{\textbf{0.54}} $\pm$ 0.06 & & 0.37 $\pm$ 0.25 & 0.18 $\pm$ 0.11 \\
\bottomrule
\end{tabular}
\caption{
Average PRR ($\pm\sigma$) across $U_{\text{info}}$ metrics for each generative paradigm. 
Entropy-derived $U_{\text{info}}$ performs strongest for finetuned and instruct paradigms. For thinking models, $U_{\text{info}}$ choice has only a minor influence on PRR. 
}
\label{tab:value_average}
\end{table}

\begin{table}[]
\centering
\small
\begin{tabular}{@{}l l@{} r r r@{}}
\toprule
\textbf{Model} & \textbf{Method} & $\textbf{PRR:~~~} U_{\text{info}}$ & $U_{\text{cons}}$ & $U_{\text{combined}}$ \\
\midrule
\multirow[c]{3}{*}{\rotatebox{90}{Llama-3.1-} \rotatebox{90}{~} \rotatebox{90}{70B-Instruct}}
 & Avg. Log-Prob. & 0.59 & - & 0.60 \\
 & Entropy & \underline{\textbf{0.60}} & - & \underline{\textbf{0.61}} \\
 & Improbability & 0.58 & - & \underline{\textbf{0.61}} \\
 & Max. Log-Prob. & 0.57 & - & 0.60 \\
 & Perplexity & 0.58 & - & 0.60 \\
 \cmidrule(lr){2-2}
 & $\Sigma$ EigV. Laplacian & - & 0.57 & - \\
 \cmidrule(lr){2-2} 
 & Self-Verbalized & 0.00 & - & 0.36 \\
\midrule

\multirow[c]{3}{*}{\rotatebox{90}{Qwen3-235B-} \rotatebox{90}{~} \rotatebox{90}{A22B-Instruct}}
 & Avg. Log-Prob. & 0.53 & - & 0.54 \\
 & Entropy & \textbf{0.54} & - & \textbf{0.57} \\
 & Improbability & 0.53 & - & 0.54 \\
 & Max. Log-Prob. & 0.51 & - & 0.53 \\
 & Perplexity & 0.53 & - & 0.56 \\
 \cmidrule(lr){2-2} 
 & $\Sigma$ EigV. Laplacian & - & 0.51 & - \\
 \cmidrule(lr){2-2}
 & Self-Verbalized & 0.01 & - & 0.28 \\
\bottomrule
\end{tabular}
\caption{
PRR of information-/consistency-based, and combined UQ methods for 2 top-performing (non-thinking) generative models.
Best PRR in bold, best overall underlined.
$U_{\text{cons}}$ is computed from 5 samples per prompt at temperature 1.
}
\label{tab:comparison_llama_qwen}
\end{table}
\begin{figure}[!h]
    \includegraphics[width=\columnwidth]{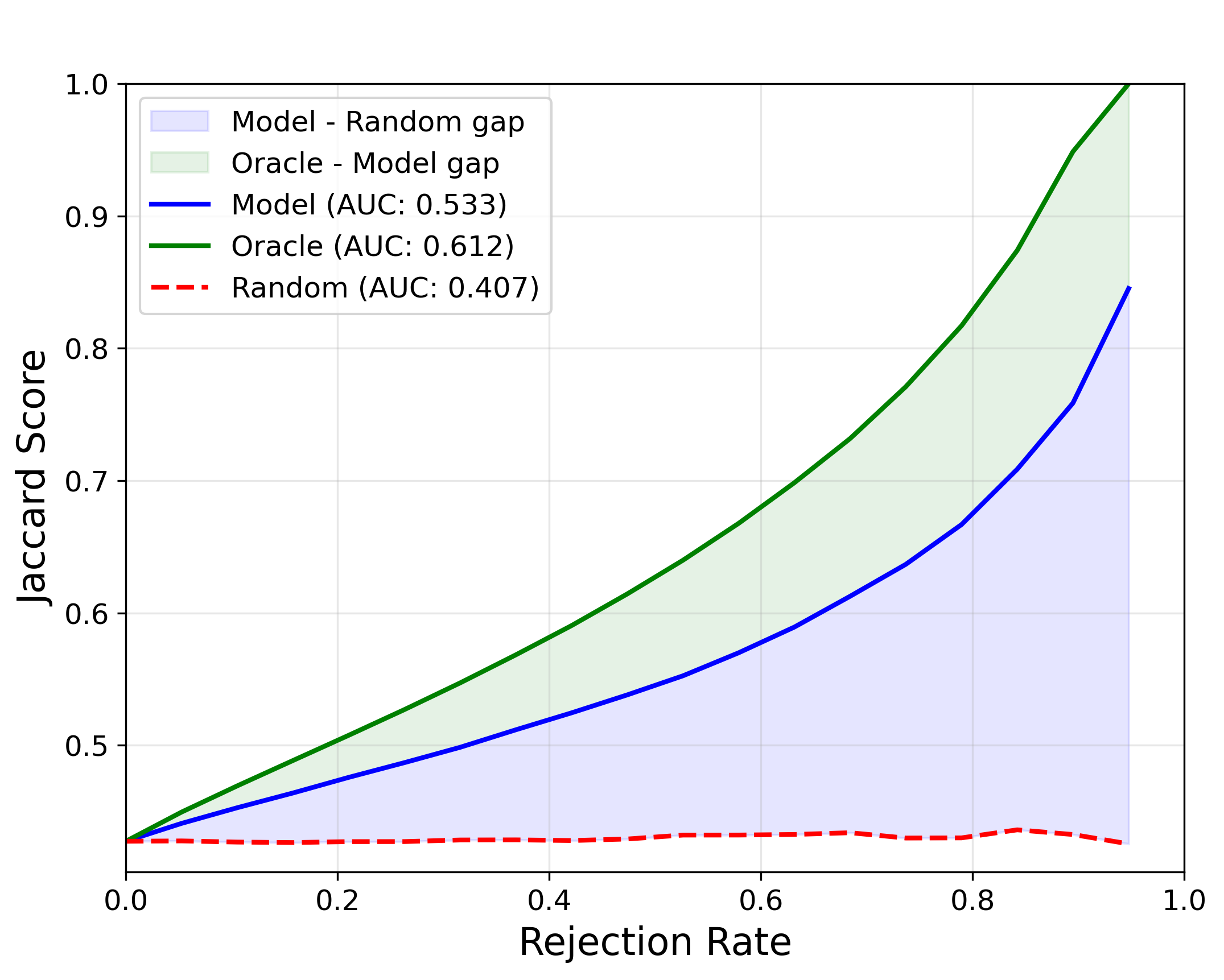}
    \caption{
    Illustration of the PRR for UQ evaluation: Rejection curves are shown for Llama-3.1-70B-Instruct under three strategies -- uncertainty-based (blue, most uncertain predictions are rejected first), oracle (green, best case, discarding predictions achieving lowest Jaccard scores first), and random (red, baseline). 
    PRR is the ratio of the model surplus to the oracle surplus: area where the uncertainty-based curve exceeds random divided by the area where the oracle exceeds random.
    }
    \label{fig:prr_illustration}
\end{figure}

\subsection{Uncertainty quantification}
In practice, reliable uncertainty estimates are what allow NLP systems to be safely deployable: a model that can signal when it does not know enables downstream pipelines to defer low-confidence predictions to human review rather than acting on them automatically, which can be more valuable than marginal accuracy gains alone.

\textbf{Fine-tuning.} 
For discriminative fine-tuning, Llama-3.2-1B-Base achieves the highest PRR (0.52) and best $\rho$, but ranks lower on $\text{ECE}{_+}$, trailing the best model in this group (Ettin-1B) by 0.04. Within the Ettin family, performance scales with model capacity, with Ettin-150M performing worst and Ettin-1B best across all metrics. This trend does not hold for the Llama family, where the smallest model (1B) achieves the best PRR (0.52) and $\rho$ (-0.42). In the generative fine-tuning setting, Llama-3.1-8B-Base attains the strongest PRR (0.63)---the highest across all paradigms---and ranks second on $\rho$. A PRR of 0.63 means that by deferring the least confident predictions, the system can eliminate a disproportionate share of errors, retaining high-precision outputs for automated processing while routing uncertain cases to human review. While still moderately underconfident, it places second on $\text{ECE}{_+}$, only 0.03 behind Llama-3.1-70B, which achieves the best $\text{ECE}{_+}$ (0.49) in this group. Notably, generative fine-tuning yields the best median PRR, $\rho$, and $\text{ECE}{_+}$ across all paradigms.

\textbf{Prompting.}
Within instruct prompting, Llama-3.1-70B achieves the strongest PRR, while Qwen3-235B-Instruct attains the best $\rho$ and $\text{ECE}{_+}$. 
Performance varies strongly across models in this paradigm ($\sigma$ of 0.15 for PRR and 0.23 for $\rho$), with some severely underperforming (e.g., Llama-3.1-8B with PRR 0.20, Kimi-K2 with PRR 0.28). Models with low PRR often exhibit higher $\rho$, consistent with expectations. $\text{ECE}{_+}$ also spans a wide range, from severe underconfidence (0.97, Kimi-K2-Instruct) to substantially better calibration (0.56, Qwen3-235B-Instruct). For prompted thinking models, PRR and $\rho$ degrade substantially, with median PRR around 0.23--roughly half of fine-tuned or instruct-prompted models. 
Several models perform only marginally above chance (PRR 0.05--0.08), and the strongest reaches 0.34 (Qwen3-235B-Thinking). 
Poor alignment between uncertainty and correctness is indicated by $\rho$ near zero or slightly negative. For applications where abstention decisions carry real costs---such as routing a case to a domain expert---this near-random uncertainty ordering makes thinking models unsuitable without further calibration, as their confidence scores provide little actionable signal for setting abstention thresholds. In contrast, this paradigm yields the best overall calibration, with an $\text{ECE}{_+}$ of 0.45 (Qwen3-235B-Thinking), 
outperforming instruct prompting and approaching fine-tuned models.

\textbf{Overall.} Generative fine-tuned models achieve the strongest UQ performance, with the best median PRR as well as $\rho$ (Mann-Whitney U test; Gen. FT group vs. all others combined; $p\le0.05$) and $\text{ECE}{_+}$, and the lowest PRR variability ($\sigma \pm 0.03$). Prompted instruct models rank second in PRR but show substantial variability across models, with several failing to provide reliable uncertainty estimates. Discriminative fine-tuning performs slightly worse in PRR but yields more stable and consistent results (Levene's test, comparing the variance of PRR; $p\le0.05$). Across all paradigms, models are markedly underconfident; even the best global $\text{ECE}{_+}$ of 0.45 indicates substantial underconfidence. 
Systematic underconfidence of this magnitude means that users relying on verbalized confidence scores would consistently perceive the model as less certain than it is, potentially over-triggering manual review and reducing the efficiency gains that selective prediction is intended to provide. Detailed $\text{ECE}{_+}$ results across label categories are shown in \ref{tab:ece_across_categories}. In contrast to fine-tuned models, thinking models fail to meaningfully quantify uncertainty in our setup. Overall, fine-tuned models---generative or discriminative---in our findings are the most reliable choice for practitioners building systems that rely on selective prediction, where the model must decide when to act autonomously versus escalate to a human reviewer. Prompted instruct models can be effective with careful model selection, while current thinking models appear unreliable and require further study. Substantial headroom for improving both selective prediction performance and calibration remains across all paradigms.

\textbf{Comparison of Uncertainty Methods.}

As discussed in Section~\ref{sec:uncertainty_quantification_methods}, uncertainty for generative models can be estimated using token-based scores ($U_{\text{info}}$), consistency-based scores ($U_{\text{cons}}$), self-verbalized uncertainty, and their combination ($U_{\text{combined}}$). Table~\ref{tab:value_average} reports average PRR and standard deviation across paradigms for different $U_{\text{info}}$ variants. Across all settings, entropy over the predicted token and its alternatives performs best. While differences among $U_{\text{info}}$ variants are moderate, we recommend entropy-based $U_{\text{info}}$ due to its strong theoretical grounding and semantic comparability to uncertainty estimates in discriminative models. Table~\ref{tab:comparison_llama_qwen} compares $U_{\text{info}}$, $U_{\text{cons}}$, self-verbalized uncertainty, and their combinations for Llama-3.1-70B-Instruct and Qwen-235B-A22B-Instruct. $U_{\text{cons}}$ provides reasonable uncertainty estimates but underperforms entropy-based $U_{\text{info}}$ by 0.03 for both models. Consistent with~\cite{lin2024generating}, combining $U_{\text{info}}$ and $U_{\text{cons}}$ yields marginal improvements over entropy-based $U_{\text{info}}$ alone, while substantially increasing computational cost due to multiple forward passes. 
In a deployment context, this cost-benefit tradeoff favors entropy-based $U_{\text{info}}$ unless computational resources are abundant and even marginal uncertainty gains translate into meaningful reductions in costly human review.
Self-verbalized uncertainty performs poorly for both models and should therefore be avoided.

\section{Conclusion}
\label{sec:discussion}

We introduce \texttt{MADE}, a living, contamination-free benchmark for MLTC, and evaluate a wide range of models for predictive performance and UQ capabilities. Our results show that task-specific discriminative models achieve state-of-the-art predictive performance and competitive UQ while remaining efficient and fully controllable. Generative fine-tuning provides modest gains for underrepresented classes but enhances UQ significantly. In contrast, prompted non-thinking models exhibit highly variable predictive and UQ capabilities, highlighting the benefits of fine-tuning. Thinking models show improved prediction on the tail, but consistently fail for UQ. We observe a persistent head--tail performance trade-off across all paradigms. Fine-tuned discriminative and generative models are notably more consistent in both predictive and uncertainty performance (lower variance) than prompted instruct and thinking models. Token-entropy-based UQ emerges as the most effective UQ mechanism, while self-verbalized uncertainty performs poorly.
As a living benchmark, \texttt{MADE} will continue to provide contamination-free evaluation as the FDA releases new quarterly reports.

Overall, MLTC remains challenging even for state-of-the-art models, with class imbalance as a primary limiting factor. 
Unlike established benchmarks where performance has largely plateaued, the best-performing model on our dataset achieves only a 54\% Macro F1 score. This gap indicates that the task remains far from solved and that the benchmark provides substantial room for differentiation between models, including future advances and UQ methods. We believe this demonstrates that, while incremental, our dataset constitutes a concrete and useful step toward mitigating benchmark saturation rather than a claim of fully addressing it.
Promising research directions for the scientific community include investigating why generative fine-tuning is beneficial for UQ, why UQ fails for thinking models, and training new reasoning models on the dataset using reinforcement learning with verifiable rewards. Further work could explore how UQ and predictive performance on ET classes can be improved for smaller encoder models. Achieving this could position them as a superior choice by combining strong predictive performance with reliable UQ, practical flexibility, and far lower computational costs compared to LLMs. 

\section{Limitations}

UQ for discriminative models can flag both positive and negative predictions where the model is uncertain, allowing human review. 
In contrast, this is not feasible for LLMs in the current setting as
flagging negative predictions would require prompting the model to explicitly enumerate all negative classes, after which UQ could be computed for those classes. 
Given the large number of classes, such generation would significantly increase computational costs, making it prohibitively expensive--an inherent drawback of LLMs and a relative advantage of discriminative models.

We jointly elicit both the classification and the model's self-verbalized uncertainty within a single prompt (see Appendix~\ref{subsec:SELF_VERBALIZED_PROMPT}). An alternative design would decouple these steps, first obtaining the model's prediction and then, in a separate prompt, eliciting its confidence or uncertainty estimate, e.g., via the $P(True)$ protocol of ~\citealp{kadavath2022languagemodelsmostlyknow}. While such a two-stage procedure may yield different or more reliable uncertainty estimates, it also incurs additional computational cost. Systematically comparing single- and multi-prompt elicitation strategies--both in terms of UQ and resource usage--remains an open direction for future work.

Loss functions, predictive performance metrics (e.g. macro F1), and UQ metrics like $U_{\text{info}}$ for discriminative models typically assume label independence.
However, in the present task, labels exhibit dependencies due to the hierarchy.
Prior work on hierarchical loss functions \cite[e.g.,][]{kim2024hierarchy} and hierarchical variants of the F1 metric \cite[e.g.,][]{kiritchenko2006learning, plaud2024revisiting, lin2025hierarchical} seeks to address these dependencies.
In this study, we provide strong comprehensive baselines using standard metrics, and conduct preliminary experiments with hierarchical losses but more extensive work needs to be done in future work.

To capture variability, we repeat generations for two models, computing $U_{\text{cons}}$ from five samples per prompt at temperature $1$ (see Table~\ref{tab:comparison_llama_qwen}).
We also report descriptive statistics where feasible (see Tables~\ref{tab:comparison_discriminative_generative_prompt} and \ref{tab:value_average}).
However, conducting multiple full training iterations (disc.~and gen.~finetuning) or inference runs (prompting) to capture performance variability would be computationally prohibitive.

Our analysis relies on a single newly introduced MLTC dataset, which limits the scope of generalization. 
We acknowledge that a single benchmark can only provide an incremental contribution and that the resulting observations may be specific to the characteristics of this dataset. Consequently, it cannot yield complete answers to the practical questions outlined in the introduction, but should instead be viewed as a starting point for broader, multi-dataset analyses.
Future work could extend this approach to further unsaturated 
datasets to validate the findings across diverse domains. 
Furthermore, the dataset is restricted to English, which may constrain applicability to multilingual settings.

\section{Ethical considerations}

We introduce this dataset of adverse event reports primarily as a MLTC benchmark and its intended use is research.
The data and associated methods may appear useful in real-world contexts, where models may help draft incident reports (in hospitals or industry) or support surveilling medical devices (in regulatory agencies).
Models could automatically label reports and check human-annotated labels.
Automation could speed up the prioritization of the large volume of adverse event reports, leading to faster follow-up on defective devices.

However, the reliability of the dataset's reports and labeling is not guaranteed (e.g., they are not extensively validated; see Appendix~\ref{subsec:data_availability}).
There is a possible bias because some of the labels originally come from manufacturers with potential vested interests.
Reports appear to be labeled by single annotators, making inter-annotator variability unknown.
Original reports have been pre-processed (e.g., partially anonymized) by the FDA. 
It is unclear how models trained on pre-processed reports behave when faced with actual raw reports in real-world applications.
Besides, model deployment would entail risks such as:
(1)~Automation bias: Over-reliance on model outputs, leading to diminished human oversight and possible reductions in expert staff under cost-saving arguments.
(2)~Missed anomalies: Models may fail to detect outlier reports of less common but severe events, or reports written in unconventional ways.
(3)~Adversarial incentives: Reporters might (un-/intentionally) phrase reports to influence automated decisions.
Given these limitations and risks, we advise against using this dataset or the associated methods for real-world reporting, compliance, or regulatory applications.
Instead, the dataset should be considered a benchmarking resource for MLTC research.


\section{Acknowledgements}
This work was supported by the Senate of Berlin 
and the European Union's Digital Europe programme under grant agreement No.~101100700 (TEF-Health).


\bibliography{custom}

\appendix
\setcounter{table}{0}
\renewcommand{\thetable}{A.\arabic{table}}
\clearpage

\setcounter{figure}{0}
\renewcommand{\thefigure}{A.\arabic{figure}}

\section{Appendix}

\subsection{Data availability}
\label{subsec:data_availability}

\textbf{Origin of the data:} 
We pre-process data that are made available by the U.S.~Food and Drug Administration\footnote{ \url{https://open.fda.gov/data/downloads/} $\rightarrow$ `Medical Device Event')} with a `Creative Commons CC0 1.0 Universal dedication' license\footnote{\url{http://creativecommons.org/publicdomain/zero/1.0/legalcode}} under the terms of service expressed by openFDA\footnote{\url{https://open.fda.gov/terms/}} 
and the data licensing version from 2014 \footnote{\url{https://open.fda.gov/license/}}. 
The reports' sources may come from manufacturers, user facilities, distributors, and voluntary sources (such as patients and physicians) as mentioned by openFDA\footnote{\url{https://open.fda.gov/apis/device/event/}}.
FDA does \underline{not} endorse this article.

\textbf{Privacy:} It is stated\footnote{\url{https://open.fda.gov/apis/}} that \textit{`openFDA [...] does not contain data with Personally Identifiable Information about patients or other sensitive information'}.
Information about the \textbf{`Responsible use of the data'} 
and a \textbf{`Disclaimer'} 
are given by openFDA.
(\textit{`Adverse event reports submitted to FDA do not undergo extensive validation or verification. Therefore, a causal relationship cannot be established between product and reactions listed in a report. While a suspected relationship may exist, it is not medically validated and should not be the sole source of information for clinical decision making or other assumptions about the safety or efficacy of a product. Additionally, it is important to remember that adverse event reports represent a small percentage of total usage numbers of a product. Common products may have a higher number of adverse events due to the higher total number of people using the product. In recent years the FDA has undertaken efforts to increase collection of adverse events. Increases in the total number of adverse events is likely caused by improved reporting.'; 
`Although MDRs are a valuable source of information, this passive surveillance system has limitations, including the potential submission of incomplete, inaccurate, untimely, unverified, or biased data. In addition, the incidence or prevalence of an event cannot be determined from this reporting system alone due to potential under-reporting of events and lack of information about frequency of device use. Because of this, MDRs comprise only one of the FDA's several important postmarket surveillance data sources.'}).

\textbf{Reproducibility:}
We publish\footnote{\url{https://github.com/raunak-agarwal/made-benchmark}} the code repository for data preprocessing and benchmark.
We make pre-processed dataset splits available\footnote{\url{https://huggingface.co/datasets/ragarwal/MADE-Multilabel-Benchmark}}. 

\subsection{Model knowledge cutoff dates}
\label{subsec:knowledge_cutoff_dates}

FDA publishes reports of adverse events involving medical devices \textit{after} each quarter through openFDA.
Thus, reports from the first quarter (July-September 2024) of our test data period (July 2024 to June 2025; see Section~\ref{sec:data}) were published in October 2024 or later.
Consequently, knowledge cutoff dates for model pre-training until including September 2024 are expected to avoid potential test data leakage (if these data were included in the pre-training dataset of a model).
Our test set should contain only data released after the knowledge cutoff dates for Llama 3.1-3 (December 2023), GPT 4.1 (June 2024), gpt-oss-120b (June 2024), and GPT-5 (September 2024).
Ettin \citep{weller2025seq} was pre-trained including on the `DOLMino mix 1124' dataset \citep[][which mentions data from September 2024]{olmo20252olmo2furious}, among other data sources. 

We have not found official information on the knowledge cutoff dates for DeepSeek-R1, Qwen3, and Kimi K2 (released in January, April, and July 2025, respectively).
Llama-3.3-Nemotron-49B-v1.5 used post-training data released in July 2025.

\subsection{Computing infrastructure}
\label{subsec:computing_infrastructure}

The computing infrastructure used included two Nvidia A100 (40GB) and up to eight Nvidia H200 and AMD Instinct MI300X accelerators.

\subsection{Methods for uncertainty quantification}
\label{sec:uncertainty_quantification_methods_appendix}

\textbf{Uncertainty for discriminative models} is quantified from the output probability vector of the MLTC model. 
We treat the
classes
as independent and
compute binary entropy per 
class:
\[\mathbb{H}(\pi_c) = - \big[ \pi_c \log(\pi_c) + (1-\pi_c)\log(1-\pi_c) \big],\] 
where $\pi_c$ is the predicted probability 
to be an instance
(label 1) of class $c$. The information-based uncertainty for a sample is the vector of entropies across 
classes:
\[
U_{\text{info}} = \big( \mathbb{H}(\pi_1), \mathbb{H}(\pi_2), \dots, \mathbb{H}(\pi_C) \big).
\]

\textbf{Uncertainty for generative models} is quantified with two complementary scores: information-based $U_{\text{info}}$ and consistency-based $U_{\text{cons}}$.
Multiplying information- and consistency-based scores results in the combined uncertainty score:
$U_{\text{combined}} = U_{\text{info}} \cdot U_{\text{cons}}$.
We compute $U_{\text{info}}$ with these 
metrics: 

\begin{itemize}
    
    \item Average Log-Probability of Tokens:
    $\text{Avg}(\pi) = -\frac{1}{L_i} \sum_j \log(\pi_{ij})$

    \item Entropy of Token Distribution:
    $H_{ij} = - \sum_{w \in D} \pi_{ij}(w) \log \pi_{ij}(w)$

    \item $\text{Improbability} = 1 - \prod_j \pi_{ij}$

    \item Maximum Log-Probability of Tokens:
    $\text{Max}(\pi) = \max_j \left(-\log(\pi_{ij})\right)$

    \item $\text{Perplexity}(\pi) = \exp\left( -\frac{1}{L_i} \sum_j \log(\pi_{ij}) \right)$
        
\end{itemize}

The consistency-based uncertainty score $U_{\text{cons}}$ is computed using the Laplacian of a graph derived from multiple stochastic forward passes \citep{lin2024generating}. 
We perform $n=5$ stochastic forward passes with temperature $t=1$, calculate the pairwise Jaccard similarity $W$ between the predictions, and compute the normalized Laplacian $L$ as $L = I - D^{-1/2} W D^{-1/2}$, where $I$ is the identity matrix and $D$ is the degree matrix ($D = \text{diag}\left(\sum_j W_{ij}\right)$).
The consistency-based uncertainty score is then $U_{\text{cons}} = \sum_{k} \max(0, 1 - \lambda_k)$, with $\lambda_k$ denoting the eigenvalues of $L$. 

Finally, we prompt the model to self-verbalize its confidence $C_i \in [0,1]$ for each prediction. High confidence corresponds to low uncertainty. Self-verbalized uncertainty is, thus,
$U_{\text{self}} = 1 - C_i$.


\subsection{Evaluation of uncertainty scores}
\label{sec:uncertainty_quantification_evaluation_appendix}

With selective prediction, we evaluate the quality of uncertainty scores (either $U_{\text{info}}$ or $U_{\text{combined}}$). Outputs of a model are rejected according to their sorted uncertainty scores. We assess whether these rejections improve the overall predictive performance.

\textbf{Prediction rejection rate} (PRR; illustrated in Figure~\ref{fig:prr_illustration}) evaluates the effectiveness of an uncertainty scoring in prioritizing unreliable predictions:
\[
\text{PRR} = \frac{\text{AUC}_{\text{uncertainty}} - \text{AUC}_{\text{random}}}{\text{AUC}_{\text{oracle}} - \text{AUC}_{\text{random}}},
\]
where
$\text{AUC}_{\text{uncertainty}}$ is the area under the curve (AUC) obtained by sorting the samples by uncertainty $U$ (highest first), iteratively rejecting the top $N\%$ of the samples, and computing the average Jaccard score after each rejection threshold.
$\text{AUC}_{\text{oracle}}$ is the AUC obtained by sorting samples by the Jaccard score (lowest score first) and iterative rejection -- this gives us a clear upper-bound to compare against for a given rejection threshold.
$\text{AUC}_{\text{random}}$ is the AUC obtained by rejecting $N\%$ random samples at each step.
A PRR close to one indicates that the uncertainty scoring is nearly as good as the oracle, while a PRR near zero indicates that the scoring is not better than random.

\textbf{Spearman correlation ($\rho$) of uncertainty with per-sample correctness.}
We evaluate binary correctness (0/1) for each example-label pair, compute Spearman's $\rho$ between $U$ and correctness separately for each label, and report the average $\rho$ across all labels. A large negative correlation indicates that higher uncertainty coincides with lower prediction accuracy, confirming that $U$ effectively flags unreliable predictions.

\textbf{Positive class expected calibration error (ECE$_{+}$).}
To assess calibration specifically on the subset of samples where a label is positively present, we compute the positive class expected calibration error. For a given class $c$, we restrict evaluation to the positive instances of that class and bin their predicted confidences into $M$ bins $b_m$. Following the formula of standard ECE, the discrepancy between empirical accuracy and predicted confidence is then
\[
\text{ECE}{_+}_c = \frac{1}{N_c^{+}} \sum_{m=1}^{M} |b_m| \, \bigl| \operatorname{acc}(b_m)^{+} - \operatorname{conf}(b_m)^{+} \bigr|,
\]
where $N_c^{+}$ is the number of samples for which class $c$ is a true label, $\operatorname{acc}(b_m)^{+}$ is the fraction of positives in bin $m$, and $\operatorname{conf}(b_m)^{+}$ is the average predicted confidence for class $c$ in that bin. Since we only consider samples where the ground truth label is present, $\operatorname{acc}(b_m)^{+} = 1$ for all bins, and the formula simplifies to
\[
\text{ECE}{_+}_c = \frac{1}{N_c^+} \sum_{m=1}^{M} |b_m| \, \bigl( 1 - \operatorname{conf}(b_m)^+ \bigr)
\]

as accuracy does not vary across bins, this further simplifies to

\[
\text{ECE}{_+}_c = 1 - \frac{1}{N_c^{+}} \sum_{i=1}^{N_c^{+}} p_i^{c},
\]

with $p_i$ being the predicted probability of class $c$ being present in a sample.
Positive class expected calibration error therefore measures average underconfidence of the model on true positive instances, and semantically captures  how well the model's confidence aligns with correctness conditional on the class being present. The total score $\text{ECE}{_+}$ is derived by averaging over all binary class scores

\[
\text{ECE}{_+} = \frac{1}{|\mathcal{C}|} \sum_{c \in \mathcal{C}} \text{ECE}{_+}_c
\]

where $\mathcal{C}$ is the set of all classes.

\subsection{Additional results}
\label{subsec:additional_results}

\begin{figure}[h!]
\includegraphics[width=\columnwidth]{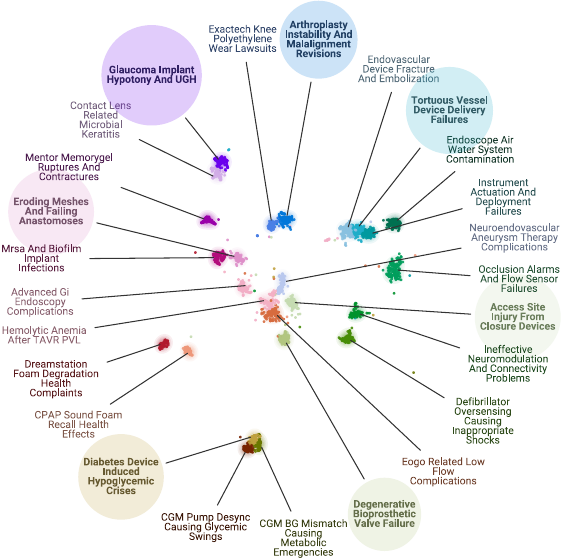}  
\caption{The 25 most frequent topics in \dataset reports (truncated test set) are identified with UMAP~\citep{mcinnes2020umapuniformmanifoldapproximation} and K-means clustering~\citep{lloyd1982least} and named with GPT-5.} 
\label{fig:topics}  
\end{figure}

Figure~\ref{fig:topics} gives an overview of the topics reported in the truncated test set of the \dataset dataset.

Table \ref{tab:bce_vs_hydra} compares a hierarchical loss \citep[HYDRA,][]{karl-scherp-2025-hydra} with standard binary cross-entropy for Ettin models. Instead of training a single classification head over the entire label space, HYDRA partitions labels by their hierarchy level and assigns a dedicated classification head to each level. Despite this, the differences between the two methods are negligible, and performance even regresses for the largest model.

\begin{table*}[h!]
\centering
\footnotesize
\begin{tabular}{@{}llccccccccc@{}}
\toprule
\textbf{Model} & \textbf{Loss} & \multicolumn{5}{c}{\textbf{Macro F1 $\uparrow$}} & \textbf{J $\uparrow$} & \textbf{PRR $\uparrow$} & $\boldsymbol{\rho} \downarrow$ & \textbf{ECE}$_{+} \downarrow$\\
\cmidrule(lr){3-7}
& & Overall & Head & Medium & Tail & ET & & & & \\
\midrule
\multirow{2}{*}{Ettin-150m-Encoder} 
  & BCE   & 0.46 & 0.68 & 0.56 & 0.44 & 0.07 & 0.55 & 0.38 & -0.30 & 0.64 \\
  & HYDRA & 0.47 & 0.69 & 0.55 & 0.44 & 0.12 & 0.56 & 0.42 & -0.41 & 0.63 \\
\midrule
\multirow{2}{*}{Ettin-400m-Encoder} 
  & BCE   & 0.51 & 0.72 & 0.61 & 0.50 & 0.12 & 0.58 & 0.44 & -0.36 & 0.59 \\
  & HYDRA & 0.49 & 0.70 & 0.57 & 0.46 & 0.14 & 0.57 & 0.45 & -0.42 & 0.61 \\
\midrule
\multirow{2}{*}{Ettin-1B-Encoder} 
  & BCE   & 0.53 & 0.73 & 0.63 & 0.51 & 0.13 & 0.61 & 0.46 & -0.40 & 0.56 \\
  & HYDRA & 0.48 & 0.70 & 0.58 & 0.44 & 0.13 & 0.57 & 0.50 & -0.43 & 0.53 \\
\bottomrule
\end{tabular}
\caption{Comparison of BCE vs HYDRA loss across Ettin encoder models.}
\label{tab:bce_vs_hydra}
\end{table*}


Full vs. parameter-efficient generative fine-tuning using LoRA are compared in Table~\ref{tab:comparison_lora_vs_full_finetuning}. Full fine-tuning improves macro-F1 by 0.01 for Llama models (1B, 3B). Gains are greater for Ettin models (400M, 1B), at 0.12 and 0.09, respectively.
Table~\ref{tab:comparison_base_vs_instruct} compares instruction-tuned models with base models. Instruction-tuned Llama models (1B, 3B) achieve macro-F1 0.04--0.05 higher than the corresponding base variants.

\begin{table*}
\centering
\footnotesize

\begin{tabular}{@{}l c c c c c c | c c c@{}}
\toprule
\textbf{Fine-tuning}/model & \multicolumn{5}{@{}c@{}}{\textbf{Macro F1 $\uparrow$}} & \textbf{J $\uparrow$} & $\textbf{PRR} \uparrow$ & \textbf{$\boldsymbol{\rho} \downarrow$} & {\centering \textbf{ECE}$_{+} \downarrow$} \\
\cmidrule(lr){2-6}
& Overall & Head & Medium & Tail & ET & & & &\\
\textit{Number of classes $\rightarrow$} & \textit{1154} & \textit{144} & \textit{481} & \textit{348} & \textit{181} & & & & \\
\midrule

\multicolumn{10}{@{}l@{}}{\textbf{Full}} \\
Llama-3.2-3B-Instruct & \underline{\textbf{0.49}} & \underline{\textbf{0.68}} & \underline{\textbf{0.57}} & \underline{\textbf{0.48}} & \underline{\textbf{0.14}} & \underline{\textbf{0.58}} & \textbf{0.59} & \textbf{-0.45} & \underline{\textbf{0.55}} \\
Llama-3.2-1B-Instruct & 0.47 & 0.67 & 0.56 & 0.45 & 0.12 & 0.57 & 0.58 & -0.43 & 0.57 \\
Ettin-1B-Decoder & 0.47 & 0.67 & 0.56 & 0.46 & 0.10 & 0.57 & 0.56 & -0.43 & 0.57 \\
Ettin-400M-Decoder & 0.44 & 0.66 & 0.54 & 0.42 & 0.07 & 0.55 & 0.56 & -0.44 & 0.60 \\
\midrule

\multicolumn{10}{@{}l@{}}{\textbf{Parameter-efficient}} \\
Llama-3.2-3B-Instruct & \textbf{0.48} & \textbf{0.67} & \underline{\textbf{0.57}} & \textbf{0.47} & \textbf{0.11} & \underline{\textbf{0.58}} & 0.58 & \underline{\textbf{-0.46}} & \textbf{0.56} \\
Llama-3.2-1B-Instruct & 0.45 & 0.66 & 0.54 & 0.44 & 0.08 & 0.56 & \underline{\textbf{0.60}} & -0.45 & 0.59 \\
Ettin-1B-Decoder & 0.38 & 0.63 & 0.48 & 0.33 & 0.014 & 0.52 &  0.58 & -0.44 & 0.67 \\
Ettin-400M-Decoder & 0.32 & 0.59 & 0.41 & 0.24 & 0.005 & 0.48 & 0.57 & -0.45 & 0.73 \\
\bottomrule
\end{tabular}

\caption{
Full vs. Parameter-Efficient Fine-Tuning (LoRA) -- Performance and UQ.
Performance metrics (macro F1, Jaccard J) and UQ measures (PRR, Spearman $\rho$, weighted ECE) are reported for both fine-tuning strategies. All models were trained in a generative setting. Macro F1 is the primary predictive metric, with results broken down by head, medium, tail, and extreme tail (ET) classes. 
PRR is reported for the best $U_{\text{info}}$ metric (see Section~\ref{sec:uncertainty_quantification_evaluation}).
While UQ results vary, full fine-tuning results often in slightly better predictive performance in comparison to the parameter-efficient version.
} 
\label{tab:comparison_lora_vs_full_finetuning}

\end{table*}

\begin{table*}
\centering
\footnotesize
\begin{tabular}{@{}l c c c c c c | c c c@{}}
\toprule

\textbf{Type}/model & \multicolumn{5}{@{}c@{}}{\textbf{Macro F1 $\uparrow$}} & \textbf{J $\uparrow$} & $\textbf{PRR} \uparrow$ & \textbf{$\boldsymbol{\rho} \downarrow$} & {\centering \textbf{ECE}$_{+} \downarrow$} \\
\cmidrule(lr){2-6}
& Overall & Head & Medium & Tail & ET & & & & \\
\textit{Number of classes $\rightarrow$} & \textit{1154} & \textit{144} & \textit{481} & \textit{348} & \textit{181} & & & & \\
\midrule

\multicolumn{10}{@{}l@{}}{\textbf{Instruction-tuned}} \\
Llama-3.2-3B-Instruct & \underline{\textbf{0.49}} & \underline{\textbf{0.68}} & \underline{\textbf{0.57}} & \underline{\textbf{0.48}} & \underline{\textbf{0.14}} & \underline{\textbf{0.58}} & \textbf{0.59} & \textbf{-0.45} & \underline{\textbf{0.56}} \\
Llama-3.2-1B-Instruct & 0.47 & 0.67 & 0.56 & 0.45 & 0.12 & 0.57 & 0.58 & -0.43 & 0.59 \\
\midrule

\multicolumn{10}{@{}l@{}}{\textbf{Base}} \\
Llama-3.2-3B-Base & \textbf{0.48} & \textbf{0.67} & \underline{\textbf{0.57}} & \textbf{0.46} & \textbf{0.12} & \underline{\textbf{0.58}} & \underline{\textbf{0.70}} & \underline{\textbf{-0.46}} & \textbf{0.59} \\
Llama-3.2-1B-Base & 0.43 & 0.63 & 0.52 & 0.39 & 0.10 & 0.45 & 0.53 & -0.44 & 0.63 \\

\bottomrule
\end{tabular}

\caption{
Instruction-Tuned vs. Base Models: Performance metrics (macro F1, J) and UQ measures (PRR, $\rho$, weighted ECE) are reported for both model types, which were fine-tuned using generative objectives. Instruction-tuned models achieve slightly higher predictive performance more often, whereas the 3B-Base model excels in UQ (highest PRR, lowest $\rho$ and ECE).
} 
\label{tab:comparison_base_vs_instruct}

\end{table*}

\begin{table*}
\centering
\footnotesize
\begin{tabular}{@{}l c c c c@{}}
\toprule

\textbf{Paradigm}/model & \multicolumn{4}{@{}c@{}}{\centering \textbf{ECE}$_{+} \downarrow$} \\
\cmidrule(lr){2-5}
& Head & Medium & Tail & ET \\
\textit{Number of classes $\rightarrow$} & \textit{144} & \textit{481} & \textit{348} & \textit{181} \\
\midrule

\multicolumn{5}{@{}l@{}}{\textbf{Discriminative fine-tuning}} \\
Llama-3.1-8B-Base         & 0.32 & 0.48 & 0.67 & 0.86 \\
Llama-3.2-3B-Base         & 0.34 & 0.50 & 0.68 & 0.86 \\
Llama-3.2-1B-Base          & 0.37 & 0.51 & 0.67 & \textbf{0.85} \\
Ettin-1B-Encoder      & \textbf{0.31} & \textbf{0.47} & \textbf{0.65} & \textbf{0.85} \\
Ettin-400m-Encoder     & 0.33 & 0.50 & 0.68 & 0.86 \\
Ettin-150m-Encoder    & 0.38 & 0.55 & 0.74 & 0.88 \\
\midrule

\multicolumn{5}{@{}l@{}}{\textbf{Generative fine-tuning}} \\
Llama-3.1-70B-Base         & \underline{\textbf{0.27}} & \underline{\textbf{0.39}} & \textbf{0.54} & \textbf{0.87} \\
Llama-3.1-8B-Base         & 0.30 & 0.42 & 0.58 & 0.91 \\
Llama-3.2-3B-Base          & 0.37 & 0.47 & 0.61 & 0.91 \\
Llama-3.2-1B-Base              & 0.40 & 0.51 & 0.66 & 0.92 \\
Ettin-1B-Decoder             & 0.37 & 0.48 & 0.61 & 0.92 \\
Ettin-400m-Decoder            & 0.39 & 0.51 & 0.65 & 0.95 \\
\midrule

\multicolumn{5}{@{}l@{}}{\textbf{Prompting - instruct}} \\
Llama-3.1-70B-Instruct            & 0.48 & 0.62 & 0.73 & 0.91 \\
Llama-3.1-8B-Instruct             & 0.64 & 0.74 & 0.83 & 0.93 \\
Qwen3-235B-A22B-Instruct    & \textbf{0.40} & \textbf{0.51} & \textbf{0.61} & \textbf{0.75} \\
Qwen3-30B-A3B-Instruct & \textbf{0.40} & 0.53 & 0.65 & 0.83 \\
Qwen3-4B-Instruct      & 0.47 & 0.61 & 0.74 & 0.90 \\
Kimi-K2-Instruct               & 0.93 & 0.96 & 0.98 & 1.00 \\
GPT-4.1                    & 0.44 & 0.56 & 0.64 & 0.79 \\
\midrule

\multicolumn{5}{@{}l@{}}{\textbf{Prompting - thinking}} \\
Llama-3.3-Nemotron-49B-v1.5        & 0.43 & 0.54 & 0.65 & 0.79 \\
Qwen3-235B-A22B-Thinking     & \textbf{0.32} & \textbf{0.40} & \underline{\textbf{0.49}} & \underline{\textbf{0.63}} \\
Qwen3-30B-A3B-Thinking & 0.44 & 0.51 & 0.60 & 0.72 \\
Qwen3-4B-Thinking      & 0.47 & 0.57 & 0.68 & 0.81 \\
DeepSeek-R1-0528           & 0.36 & 0.45 & 0.53 & 0.67 \\
GLM-4.5-Air & 0.44 & 0.56 & 0.69 & 0.82 \\
GPT-5  & NA & NA & NA & NA \\

\bottomrule
\end{tabular}

\caption{
Positive class expected calibration errors (head, medium, tail, extreme tail) for each paradigm/model.
}
\label{tab:ece_across_categories}
\end{table*}

\subsection{Prompting setup}
\label{subsec:Prompting_setup}

\lstset{
  basicstyle=\ttfamily\small,
  numberstyle=\tiny,
  keywordstyle=\color{blue},
  commentstyle=\color{gray},
  stringstyle=\color{red},
  breaklines=true
}

\subsubsection{System prompt}
\label{subsec:SYS_PROMPT}
\begin{lstlisting}[caption={}]
SYSTEM_PROMPT = """
You are an AI assistant tasked with classifying a medical device adverse event report into one or more categories according to the FDA taxonomy. Your goal is to assign all relevant labels to the given report.

The report that needs classifying is provided within the <classification-text> tag. Along with the report, the label definitions are provided within the <labels> tag. To assist you with the task, we also include 10 "few-shot" examples in the <few-shot-examples> tag. These are past reports similar to the one you are classifying - the past reports are accompanied by their corresponding labels which were tagged by a human expert.

# RULES:

1. The taxonomy of labels is provided within the <labels> tag.
    - Labels are separated by newlines; a definition for the label is provided.
    - We are in a 3-level hierarchical multi-label classification setting - this means that when a child label (such as A040507) is selected, the parent label (A0405) and grandparent label (A04) must also be selected. Similarly, if a parent label (A0405) is selected, the grandparent label (A04) must also be selected.
    - The converse is not always true - selecting a parent (A0405) or grandparent (A04) doesn't necessarily mean selecting all its children (A040507).
    - Labels that start with "A" are Medical Device Problems.
    - Labels that start with "E" are Health Effects - Clinical Signs and Symptoms or Conditions.
    - The grandparent label (eg. A01) is the most general label, the parent label (eg. A0101) is the next most specific, and the child label (eg. A010101) is the most specific.
    - The grandparent label always has the letter "A" or "E" followed by 2 numbers, eg. A01, E01, A02, E02, etc.
    - The parent label always has the letter "A" or "E" followed by 4 numbers, eg. A0101, E0101, A0201, E0201, etc.
    - The child label always has the letter "A" or "E" followed by 6 numbers, eg. A010101, E010101, A020101, E020101, etc.

2. There are 10 "few-shot" examples included in the <few-shot-examples> tag.
    - Each example includes a report and its corresponding labels.
    - The examples included in the <few-shot-examples> tag were chosen using a K-Nearest Neighbours algorithm which picked reports similar in content to the text which needs classifying. The labels shown for these examples may or may not overlap with the labels for the report inside the <classification-text> tag. Use them as contextual guidance.

3. Your goal is to classify the text provided within the <classification-text> tag.
    - Assign all labels that are relevant.
    - You can choose multiple labels, a single label, or no labels if none apply.
    - Always use the exact label names from the label list provided in the taxonomy under the <labels> tag.  Do not invent new labels or modify existing ones.
    - Return your output as a list of labels, separated by newlines.
    - Do not include any explanations, text, or formatting outside the label list.
    - If no label applies, return an empty list.
    - Do not invent new labels.

4. Provide your output as a list of labels, each on a new line. For example:

A04
A0405
A040507
E01
E0101

# IMPORTANT
*In your final output, you must not include any extra text, explanations, or formatting outside the label list. Only return the list of labels separated by newlines.*
"""
\end{lstlisting}

\subsubsection{User prompt}
\label{subsec:SYS_PROMPT}
\begin{lstlisting}[caption={}]
USER_PROMPT = """
You are an AI assistant tasked with classifying a medical device adverse event report into one or more categories according to the FDA taxonomy. Your goal is to assign all relevant labels to the given report. The rules were provided in the system prompt. For the sake of clarity, we are repeating them here:
- Assign all the labels that are relevant to the report, but only if you are sure about it.
- You can choose multiple labels, a single label, or no labels if none apply.
- Use exact label names from the provided taxonomy. Do not invent or modify labels.
- Include parent and grandparent labels when selecting a child label.
- Do not include any explanations, text, or formatting outside the label list.
- If no labels apply, return an empty list.
- Your output should contain only the list of applicable labels, with each label on a new line. You must not include any extra text, explanations, or formatting outside the label list.
- Provide your output as a list of labels, each on a new line. 
Example output:
A04
A0405
A040507
E01
E0101

Here's how to proceed:

1. First,  familiarize yourself with the label definitions:

<labels>
A01: Patient Device Interaction Problem - Problem related to the interaction between the patient and the device.
A0101: Patient-Device Incompatibility - Problem associated with the interaction between the patient's physiology or anatomy and the device that affects the patient and/or the device.
.
.

</labels>

2. Review these few-shot examples of similar reports and their corresponding labels:

<few-shot-examples>
{EXAMPLES}
</few-shot-examples>

3. Now, carefully classify the following report:

<classification-text>
{CLASSIFICATION_TEXT}
</classification-text>"""

\end{lstlisting}

\subsubsection{Variation With Self-Verbalized Confidence}
\label{subsec:SELF_VERBALIZED_PROMPT}
\begin{lstlisting}[caption={}]
"""
...
- Assign all labels that are relevant.
- You can choose multiple labels, a single label, or no labels if none apply.
- Always use the exact label names from the label list provided in the taxonomy under the <labels> tag.  Do not invent new labels or modify existing ones.
- Return your output as a JSON dictionary with label codes as keys and confidence scores as values. Confidence scores should be floats between 0 and 1, reflecting your certainty about each prediction.
- Do not include any explanations, text, or formatting outside the JSON dictionary.
- If no label applies, return an empty JSON object: {}.
- Do not invent new labels.

Provide your output as a JSON dictionary.
For example:
{
"A04": 0.92,
"A0405": 0.74,
"A040507": 0.50,
"E01": 0.95,
"E0101": 0.32
}
# IMPORTANT
* In your final output, you must not include any extra text, explanations, or formatting outside the JSON dictionary. Only return the JSON dictionary.
"""
\end{lstlisting}

\end{document}